%% file: main.tex
\documentclass[sigconf,natbib=true,nonacm]{acmart}

\setcopyright{none}
\makeatletter
\AtBeginDocument{%
  \def\@copyrightpermission{}%
}
\makeatother

\usepackage{multirow}
\usepackage{tabularx}

\usepackage{xcolor}
\usepackage{tcolorbox}
\tcbuselibrary{skins}

\definecolor{boxframe}{RGB}{50, 50, 50}
\definecolor{boxback}{RGB}{248, 248, 250}
\definecolor{tagcolor}{RGB}{0, 0, 150}

\newtcolorbox{promptbox}[1][]{
  enhanced,
  title=\textsc{System Prompt Template},
  colframe=boxframe,
  colback=boxback,
  coltitle=white,
  fonttitle=\bfseries,
  attach boxed title to top left={xshift=4mm, yshift=-2mm},
  boxed title style={
    size=small,
    colback=boxframe,
    sharp corners=south,
    rounded corners=north,
  },
  top=4mm,
  drop shadow,
  #1
}

\begin{document}

\title{Towards Transparent RAG: Fostering Evidence Traceability in LLM Generation via Reinforcement Learning}


\author{Jingyi Ren}
\affiliation{
  \institution{DCST \& AIR, Tsinghua University}
  \country{Beijing, China}
}

\author{Yekun Xu}
\affiliation{
  \institution{DCST, Tsinghua University}
  \country{Beijing, China}
}

\author{Xiaolong Wang}
\affiliation{
  \institution{DCST, Tsinghua University}
  \country{Beijing, China}
}

\author{Weitao Li}
\affiliation{
  \institution{DCST, Tsinghua University}
  \country{Beijing, China}
}

\author{Ante Wang}
\affiliation{
  \institution{AIR, Tsinghua University}
  \country{Beijing, China}
}

\author{Weizhi Ma}
\authornote{Corresponding authors.}
\affiliation{
  \institution{AIR, Tsinghua University}
  \country{Beijing, China}
}
\email{mawz@tsinghua.edu.cn}

\author{Yang Liu}
\authornotemark[1]
\affiliation{
  \institution{DSCT \& AIR, Tsinghua University}
  \country{Beijing, China}
}
\email{liuyang2011@tsinghua.edu.cn}

\renewcommand{\shortauthors}{Preprint. Under review.}

\begin{abstract}

Retrieval-Augmented Generation (RAG) delivers substantial value in knowledge-intensive applications. However, its generated responses often lack transparent reasoning paths that trace back to source evidence from retrieved documents. This opacity not only compromises the interpretability of the output but also limits the model's ability to fully exploit the provided context.
To address this, we propose \textbf{TRACE} (\textbf{T}ransparent \textbf{RA}G with eviden\textbf{CE} tracing), a framework designed to enhance evidence traceability in Large Language Models (LLMs) through reinforcement learning (RL).
TRACE guides LLMs to produce structured outputs with explicit evidence citations by prompting and rewarding evidence relevance and proper formatting, alongside accuracy, to optimize structured traceability.
To ensure training stability with multiple reward signals, we further introduce an adaptive strategy for merging rewards and adopt a stabilized KL-divergence estimator.
Experiments on three multi-hop QA datasets using Qwen2.5-7B-Instruct and Llama-3.1-8B-Instruct show that TRACE achieves both transparent, evidence-attributed outputs and accuracy improvements of 10–30\%. The resulting performance is comparable to advanced commercial LLMs (e.g., OpenAI o1, DeepSeek-R1). Further analyses demonstrate strong generalization capabilities to unseen tasks.
Our code is publicly available now\footnote{\url{https://github.com/ren258/TRACE}}.

\end{abstract}

\maketitle

\section{Introduction}
\label{sec:introduction}

Retrieval-Augmented Generation (RAG)~\cite{lewis2020retrieval} has become a powerful paradigm for enhancing large language models (LLMs) with non-parametric knowledge~\cite{gao2023retrieval,gupta2024comprehensive,zhou2025depth}, particularly in knowledge-intensive domains such as medicine~\cite{li2024agent}, law~\cite{wiratunga2024cbr}, and finance~\cite{setty2024improving}. Beyond accuracy gains from retrieved context~\cite{yu2022retrieval}, practical deployment further requires transparency, auditability, and decision traceability, so that one can verify which evidence supports which reasoning steps and the final answer~\cite{singh2024rethinking,yeo2024interpretable}. 

Recent research emphasizes the active acquisition of information, often interleaving search and reasoning steps when initial documents are insufficient~\cite{gao2025beyond, li2025ur}. By dynamically expanding the context, these methods have achieved remarkable accuracy improvements on complex multi-hop QA tasks, particularly through Reinforcement Learning (RL) strategies like Group Relative Policy Optimization (GRPO)~\cite{shao2024deepseekmath, searchr1}. 
However, despite their strong performance, most these approaches typically \textbf{operate as black boxes} regarding evidence utilization; they do not make explicit or verifiable which specific retrieved references are actually utilized during the intermediate reasoning that leads to the final answer~\cite{research}. 
Another stream of work focuses on enhancing interpretability in a \textbf{post-generation manner}, either by appending citation tags to generated statements~\cite{webgpt, longcite} or establishing post-hoc links between generated content and retrieved materials~\cite{vericite, ceg}. 
While effective for auditability, these attribution-centric methods generally treat citation as a requirement for transparency rather than a mechanism for logic enforcement, leaving the potential of leveraging precise evidence selection to conversely enhance reasoning accuracy largely underexplored.

In this work, we investigate how transparent attribution can enhance the performance of RAG generators given retrieved results. We propose \textbf{TRACE} (\textbf{T}ransparent \textbf{RA}G with eviden\textbf{CE} tracing), a framework designed to unify reasoning accuracy and decision transparency via reinforcement learning. 
First, TRACE enforces a structured output protocol that explicitly includes the selected references, step-wise reasoning traces, and the final answer, rendering the entire evidence–reasoning–answer chain auditable. 
Second, to align the model with this objective, we implement task-specific reward functions encouraging correct answers, faithful citation usage, and well-formed decision traces; crucially, to promote holistic alignment, we introduce a bonus mechanism that assigns an amplified reward if and only if the response is perfect across all dimensions. 
Finally, we address the optimization challenges in constrained generation by substituting the default KL divergence estimator with a gradient-unbiased estimator. This modification mitigates the training instability issues identified in prior studies~\cite{tang2025few,yu2025dapo}, ensuring robust convergence even under rigid formatting constraints.

We implement TRACE on two representative open-source backbones: Qwen2.5-7B-Instruct~\cite{yang2024qwen2} and Llama3.1-8B-Instruct~\cite{grattafiori2024llama}. 
Extensive evaluations on three complex multi-hop QA benchmarks (HotpotQA~\cite{yang2018hotpotqa}, 2WikiMultiHopQA~\cite{ho2020constructing}, and MuSiQue~\cite{trivedi2022musique}), demonstrate that our method produces highly transparent outputs while delivering substantial accuracy improvements of 10--30\% over backbones. 
Notably, our results indicate that open-source models trained with TRACE exhibit reasoning capabilities approaching those of advanced commercial models such as OpenAI o1~\cite{openai2024reasoning} and DeepSeek-R1~\cite{song2025r1}. 
Moreover, generalization experiments confirm that TRACE maintains superior performance across out-of-domain (OOD) test sets and diverse real-world retrieval scenarios, including both local and web-based settings. 
Finally, comprehensive ablation studies verify the necessity and effectiveness of each proposed component within our framework.

Our main contributions are summarized as follows:
\begin{itemize}
\item We propose \textbf{TRACE}, a reinforcement learning framework that synergizes evidence attribution with reasoning accuracy. We utilize RL to optimize the joint generation of evidence traces and answers, enabling the model to leverage grounding to enhance performance.
\item We design a customized optimization strategy specifically for RAG reasoning tasks, incorporating a holistic alignment bonus to resolve gradient conflicts and a gradient-unbiased KL estimator to mitigate the training instability often encountered in complex evidence processing.
\item We validate TRACE across three multi-hop benchmarks, demonstrating that it achieves 10--30\% accuracy improvements, consistently surpassing all baselines of comparable scale and exhibiting comparable reasoning capabilities than top-tier commercial models.
\end{itemize}

\section{Related Work}

\subsection{Retrieval-Augmented Generation for Complex Reasoning}
Retrieval-Augmented Generation (RAG) ~\cite{lewis2020retrieval} fundamentally enhances LLMs by grounding generation in external evidence, thereby reducing parametric hallucinations and improving factual consistency on knowledge-intensive tasks~\cite{guu2020retrieval, gao2023retrieval, shuster2021retrieval, chen2024benchmarking}. 
However, applying RAG to complex multi-hop Question Answering (QA) tasks~\cite{yang2018hotpotqa, ho2020constructing, trivedi2022musique} presents unique challenges, as these problems require synthesizing information from multiple disparate documents to derive the correct answer~\cite{press2022measuring}. 
Unlike simple single-hop queries, multi-hop reasoning is highly sensitive to the quality of the retrieved context. 
As retrieval systems scale to include a larger number of top-ranked documents to ensure coverage, the inclusion of irrelevant or noisy context inevitably increases~\cite{petroni2020context}. 
Prior studies indicate that such noise can severely mislead the generation process, causing models to hallucinate or misinterpret relationships between entities~\cite{creswell2022selection}. 
Furthermore, when faced with long and noisy contexts, LLMs often exhibit the ``lost-in-the-middle''\cite{liu2024lost} phenomenon or simply ignore the retrieved evidence in favor of their internal priors~\cite{yoran2023making}. 
This inability to effectively filter and utilize retrieved context underscores the necessity for generation mechanisms that can explicitly trace and anchor reasoning steps to specific evidence.

To tackle the complexity of multi-hop reasoning, recent research has evolved building sophisticated systems that integrate retrieval directly into the reasoning chain. These approaches can be broadly categorized into prompt-based decomposition, SFT-based alignment, and RL-driven active retrieval.

\textit{Prompt-Based Decomposition.}
Early attempts primarily utilized the inherent capabilities of frozen LLMs to break down complex queries. Methods like Self-Ask~\cite{press2022measuring} and IRCoT~\cite{trivedi2022interleaving} guide the model to decompose multi-hop questions into simpler sub-questions, retrieving documents for each step iteratively. 
Similarly, SuRe~\cite{kim2024sure} enhances performance by summarizing retrieval results to reduce noise before generation. 
While effective without training, these inference-time strategies often suffer from high latency and are constrained by the base model's ability to handle long contexts~\cite{ma2023query}.

\textit{SFT and Synthetic Data Engineering.}
To internalize these reasoning capabilities, supervised fine-tuning (SFT) approaches have gained prominence. 
SimpleDeepSearcher~\cite{simpledeepsearcher} demonstrates that strategic data engineering which training models on automatically generated reasoning and retrieval trajectories can offer a competitive alternative to complex pipelines. 
However, SFT methods are inherently limited by the quality of static annotations and often struggle to generalize to unseen retrieval scenarios where the gold reasoning path is ambiguous.

\textit{Reinforcement Learning and Active RAG.}
The most significant advancements have come from the ``Active RAG'' paradigm, which utilizes Reinforcement Learning (RL) to empower models to autonomously interact with external knowledge. 
Foundational works like WebGPT~\cite{webgpt} used behavior cloning for web browsing, while IM-RAG~\cite{imrag} and SIM-RAG~\cite{simrag} focused on optimizing the decision of \textit{when} to retrieve, judging information sufficiency to reduce unnecessary queries.
Building on this, recent agentic RAG frameworks treat retrieval as an executable action within a Chain-of-Thought (CoT) process. 
Search-o1~\cite{searcho1} and Search-R1~\cite{searchr1} incorporate search modules optimized via outcome rewards to refine documents dynamically. 
R1-Searcher++~\cite{r1searcher++} further advances this by training the model to dynamically balance its internal parametric knowledge with external search results. 
Parallel efforts such as ReSearch~\cite{research}, CoRAG~\cite{corag}, and InstructRAG~\cite{instructrag} explicitly model the retrieval process as a multi-turn interaction, allowing the generator to reformulate queries based on partial evidence. 
To mitigate the high costs and noise associated with training these agents on real search engines, ZeroSearch~\cite{zerosearch} and O$^2$-Searcher~\cite{o2searcher} introduce simulated search environments combined with curriculum learning to robustly train retrieval agents.

Despite their sophisticated mechanisms for acquiring information and interleaving search with reasoning heavily optimize the process of searching or the final answer accuracy, but rarely incentivize the model to transparently trace which specific pieces of the retrieved information supported each step of the reasoning. 
While they may achieve high performance, they do not make explicit or verifiable which retrieved references are actually used during the reasoning that leads to the final answer, leaving the verification gap unresolved in high-stakes deployments.

\subsection{Attribution and Traceability in RAG}
To address the opacity of RAG generation, a parallel line of research focuses on ensuring that model outputs are not only accurate but also explicitly attributable to source documents.

\textit{Benchmarks and Evaluation.}
The first step towards traceability lies in rigorous evaluation. Seminal benchmarks such as ALCE~\cite{alce} have established automatic metrics to quantify citation quality, revealing that even advanced models frequently generate fluent but unsupported claims. 
Building on this, RAGAS~\cite{ragas} and ARES~\cite{ares} introduced reference-free evaluation frameworks utilizing LLMs as judges to measure faithfulness. 
In the domain of long-context generation, LongCite~\cite{longcite} proposed the LongBench-Cite benchmark to assess fine-grained citation performance. 
Webis-CrowdRAG-25~\cite{gienapp2025viability} further investigated human attribution behaviors, suggesting that verifiable generation requires assessing the tight coupling between evidence and claims.

\textit{Inference-Time and Post-Hoc Solutions.}
To improve traceability without retraining, several studies have explored inference-time interventions. 
VeriCite~\cite{vericite} adopts a multi-stage pipeline that generates an initial answer and subsequently verifies claims against retrieved documents. 
Similarly, CEG~\cite{ceg} employs a post-hoc approach where a natural language inference model detects unsupported statements and triggers regeneration. 
Taking a different angle, MIRAGE~\cite{mirage} analyzes model internals to detect salient tokens that influence the output. 
While these plug-and-play methods offer improved verifiability, they typically incur significant latency overheads and treat the lack of traceability as a symptom to be patched rather than a fundamental reasoning capability to be learned.

\textit{Training-Based Alignment and Optimization Challenges.}
Consequently, recent works aim to internalize attribution capabilities directly into the model. 
Early efforts like WebGPT~\cite{webgpt} and WebGLM~\cite{webglm} utilized imitation learning to teach models how to browse and cite web sources. 
LongCite constructed a large-scale dataset for SFT to enable precise sentence-level citations. 
APO~\cite{apo} further advanced this by modeling citation generation as a preference optimization problem.

However, a critical limitation persists in current paradigms. Most existing approaches treat citation primarily as a post-hoc annotation step, typically generating the answer or rationale first and appending references retrospectively. This strategy  failing to fully exploit the potential of precise evidence selection to conversely enhance reasoning accuracy.
In contrast, determining the relevant references before conducting the reasoning process ensures that the subsequent generation is strictly grounded in verified facts, thereby significantly improving overall performance.
Nevertheless, enforcing such a strict pre-selection protocol via reinforcement learning introduces specific optimization challenges. Imposing rigid structural constraints often leads to training instability when using standard KL divergence estimators, as they can induce sharp gradient spikes during policy updates~\cite{tang2025few, yu2025dapo}.
Our TRACE framework addresses these issues by synergizing evidence-first generation with a stabilized optimization strategy.

\section{Preliminaries}
\label{sec:preliminaries}

In this section, we provide the formal definition of the Retrieval-Augmented Generation (RAG) task and review the Group Relative Policy Optimization (GRPO) algorithm. These foundations serve as the basis for our proposed TRACE framework.

\subsection{Formalization of RAG}
\label{sec:rag_formal}

The standard RAG process involves two primary components: a retriever $\mathcal{R}$ and a probabilistic generator $\pi_\theta$ parameterized by $\theta$. Let $\mathcal{D} = \{d_1, d_2, \dots, d_M\}$ denote a large-scale knowledge corpus. Given a user query $q$, the retriever first selects a subset of relevant documents $\mathcal{C} = \{c_1, c_2, \dots, c_K\} \subset \mathcal{D}$ (where $K$ is the retrieval depth) based on semantic similarity:
\begin{equation}
    \mathcal{C} = \mathcal{R}(q, \mathcal{D}).
\end{equation}
Subsequently, the generator $\pi_\theta$ conditions on both the query and the retrieved context to produce a response output $o$:
\begin{equation}
    o \sim \pi_\theta(o \mid q, \mathcal{C}).
\end{equation}

In this work, we focus specifically on optimizing the reasoning and attribution capabilities of the generator. Consequently, we assume the retrieval results $\mathcal{C}$ are provided by an off-the-shelf retriever and remain \textit{fixed} during the training of the generator.

\begin{figure*}[t]
    \centering
    \includegraphics[width=1\textwidth,
        trim=0mm 4mm 0mm 4mm
    ]{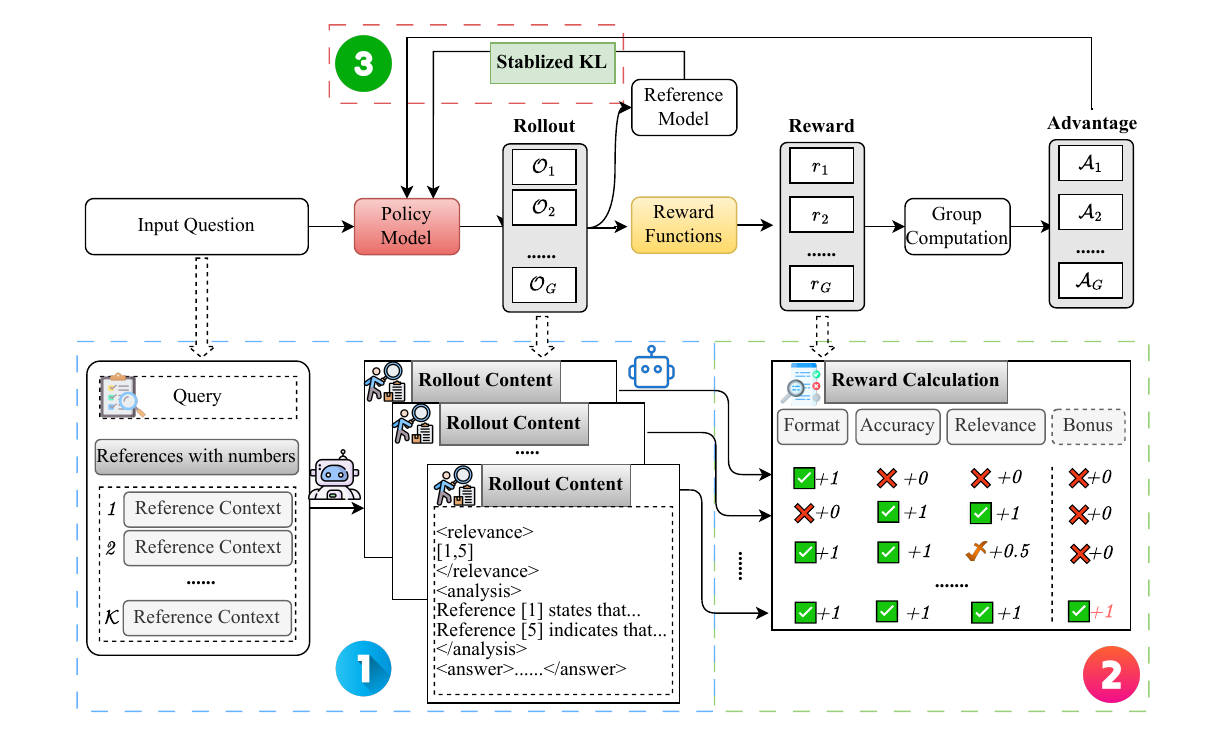} 
    \caption{
    Overview of the \textbf{TRACE} framework. The training process integrates three key components to foster transparency and reasoning: 
\textcircled{1} \textbf{Structured Protocol}, where the policy model generates a multi-part response including selected evidence, reasoning analysis, and a final answer; 
\textcircled{2} \textbf{Adaptive Reward Calculation}, where outputs are evaluated on three core dimensions—format, accuracy, and relevance. A \textbf{Bonus} mechanism is introduced to amplify the gradient signal only when all dimensions are perfectly aligned;
\textcircled{3} \textbf{Stabilized KL Divergence}, we replace the standard estimator with a stabilized estimator to mitigate KL loss.
    }
    \label{fig:pipeline}
\end{figure*}

\subsection{Reinforcement Learning with GRPO}
\label{sec:grpo_prelim}

While Supervised Fine-Tuning (SFT) is effective for basic instruction following, it often falls short in complex reasoning tasks due to the scarcity of high-quality, process-annotated data. Reinforcement Learning (RL) addresses this limitation by enabling models to \textit{self-explore} solution spaces and refine their reasoning strategies through trial and error. To instantiate this, we employ Group Relative Policy Optimization (GRPO)~\cite{shao2024deepseekmath}, which is an advanced policy optimization algorithm that improves over Proximal Policy Optimization (PPO)~\cite{schulman2017proximal} by eliminating the need for a critic model, thus reducing computational overhead. 

\paragraph{Group Advantage Estimation.}
For each query $q$, GRPO samples a group of $G$ outputs $\{o_i\}_{i=1}^G$ from the old policy $\pi_{\theta_{old}}$. The advantage $\hat{A}_i$ for the $i$-th output is derived solely from the group's reward statistics. Let $r_i$ be the total reward for output $o_i$, the advantage is computed as:
\begin{equation}
    \hat{A}_{i} = \frac{r_i - \mu_r}{\sigma_r + \delta},
\end{equation}
where $\mu_r$ and $\sigma_r$ are the mean and standard deviation of the group rewards, and $\delta$ is a small constant for numerical stability.

\paragraph{Objective Function.}
GRPO updates the policy by maximizing a surrogate objective that encourages high-advantage actions while penalizing large policy shifts. Let $\eta_{i,t}(\theta)$ denote the probability ratio between the current and old policies at token $t$:
\begin{equation}
    \eta_{i,t}(\theta) = \frac{\pi_\theta(o_{i,t} \mid q, o_{i,<t})}{\pi_{\theta_{old}}(o_{i,t} \mid q, o_{i,<t})}.
\end{equation}
The token-level clipped surrogate loss $\ell_{i,t}(\theta)$ is then defined to enforce a pessimistic bound on the policy update:
\begin{equation}
    \ell_{i,t}(\theta) = \min \left( 
        \eta_{i,t}(\theta) \hat{A}_i, \, 
        \text{clip}\left(\eta_{i,t}(\theta), 1-\varepsilon, 1+\varepsilon\right) \hat{A}_i 
    \right),
\end{equation}
where $\varepsilon$ is the clipping hyperparameter. 

The final GRPO training objective averages this surrogate loss over all groups and tokens, incorporating a KL divergence penalty to regularize the policy towards a reference model $\pi_{ref}$:
\begin{equation}
    \mathcal{J}(\theta) = \mathbb{E}_{q \sim \mathcal{D}, \{o_i\} \sim \pi_{\theta_{old}}} \left[ 
    \frac{1}{G} \sum_{i=1}^G \frac{1}{|o_i|} \sum_{t=1}^{|o_i|} 
    \left( \ell_{i,t}(\theta) - \beta \, \hat{\mathbb{D}}_{KL} \right)
    \right].
\end{equation}

\paragraph{KL Divergence Estimation.}
A critical component of the GRPO objective is the KL divergence term, which prevents the updated policy $\pi_\theta$ from deviating excessively from the reference policy $\pi_{ref}$. Formally, the exact KL divergence at token $t$ is defined as the expectation over the vocabulary $\mathcal{V}$:
\begin{equation}
    \mathbb{D}_{KL}(\pi_\theta || \pi_{ref}) = \sum_{v \in \mathcal{V}} \pi_\theta(v \mid q, o_{i,<t}) \log \frac{\pi_\theta(v \mid q, o_{i,<t})}{\pi_{ref}(v \mid q, o_{i,<t})}.
\end{equation}

However, computing this exact summation over the entire vocabulary at every step is computationally intractable. Therefore, efficient implementations approximate this value using the specific token $o_{i,t}$ sampled during the rollout. Let $\rho_t$ denote the probability ratio between the reference and current policy:
\begin{equation}
    \rho_t = \frac{\pi_{ref}(o_{i,t} \mid q, o_{i,<t})}{\pi_\theta(o_{i,t} \mid q, o_{i,<t})}.
\end{equation}
Standard implementations commonly adopt the \textbf{$k_3$ estimator}~\cite{schulman2017proximal} for this approximation:
\begin{equation}
    \hat{\mathbb{D}}_{KL}^{(k_3)} = \rho_t - \log \rho_t - 1.
\end{equation}
This estimator ensures that the penalty is minimized when $\rho_t = 1$. However, this standard estimation can exhibit instability under significant distribution shifts, motivating our proposed modifications.

\section{TRACE: Transparent RAG with Evidence Tracing}
\label{sec:trace}

We introduce \textbf{TRACE} (\textbf{T}ransparent \textbf{RA}G with eviden\textbf{CE} tracing), a RL framework designed to improve evidence traceability and reasoning stability in RAG systems. As illustrated in Figure~\ref{fig:pipeline}, the framework optimizes the generator through three components: (1) a \textbf{Structured Protocol} that enforces explicit evidence grounding and separated reasoning; (2) \textbf{Adaptive Rewards} that aligns the model with multi-dimensional objectives including format, accuracy, and relevance; and (3) a \textbf{Stabilized KL Divergence} estimator that ensures robust policy updates through gradient unbiasedness. 

\subsection{Structured Reasoning Protocol}
\label{sec:protocol}

Standard RAG generators typically output a text sequence with mixing reasoning, citation, and conclusion, which makes it difficult to verify whether the model hallucinated or correctly utilized the retrieved context. To address this, we design a strict output structure, as shown in Figure~\ref{fig:prompt}.

\begin{figure}[t]
\centering
\begin{promptbox}
\small\ttfamily

A conversation between User and Assistant. The user asks a question and gives some references. The assistant should answer the question based on the references. 

User's input will always contain:

\textcolor{tagcolor}{<question>} [The question to answer] \textcolor{tagcolor}{</question>}

\textcolor{tagcolor}{<references>} [References starting with numbers] \textcolor{tagcolor}{</references>}

\vspace{2mm}
Assistant's response must contain EXACTLY three sections:

\textcolor{tagcolor}{<relevance>}
[List ONLY reference numbers that provide useful information in square brackets, e.g. [1,5]]
\textcolor{tagcolor}{</relevance>}

\textcolor{tagcolor}{<analysis>}
[Combine information from relevant references to build the answer. Explicitly mention which references support each claim]
\textcolor{tagcolor}{</analysis>}

\textcolor{tagcolor}{<answer>}
[Answer with ONLY a short phrase or single word. No explanations]
\textcolor{tagcolor}{</answer>}

\vspace{2mm}
\textbf{**User**:}\\
\textcolor{tagcolor}{<question>} \{question\} \textcolor{tagcolor}{</question>}

\textcolor{tagcolor}{<references>} \{references\} \textcolor{tagcolor}{</references>}
\end{promptbox}
\caption{The structured prompt template used in TRACE.}
\label{fig:prompt}
\end{figure}

By conditioning the generation on this schema, we transform the generator's output space from unstructured text into a structured tuple $(\mathcal{I}, \mathcal{Z}, \mathcal{O})$. Specifically, the prompt maps the raw generation into three distinct components:

\begin{itemize}
    \item \textbf{Evidence Set $\mathcal{I}$ (via \texttt{<relevance>})}: A discrete set of indices $\mathcal{I} \subseteq \{1, \dots, K\}$ indicating exactly which documents from the retrieved context $\mathcal{C}$ are utilized. 
    \item \textbf{Reasoning Trace $\mathcal{Z}$ (via \texttt{<analysis>})}: A natural language rationale that synthesizes information from the selected evidence to deduce the answer. 
    \item \textbf{Final Answer $\mathcal{O}$ (via \texttt{<answer>})}: A concise conclusion derived directly from the reasoning trace. 
\end{itemize}

A key design choice in Figure~\ref{fig:prompt} is mandating the <relevance> block to appear first. This enforces retrieval-aware planning, compelling the model to decide what information is necessary before deciding how to use it. This structure not only mitigates post-hoc rationalization but also enables the granular reward calculations described in the next section.

\subsection{Adaptive Reward Calculation}
\label{sec:rewards}

In the standard GRPO framework, the total reward $r_i$ for an output $o_i$ is typically computed as a weighted sum of $M$ specific reward components: $r_i = \sum_{j=1}^M w_j R_j(o_i)$. While naive implementations often rely solely on format constraints and final answer accuracy, fostering traceable reasoning requires a more granular evaluation. Therefore, we design four task-specific reward functions to guide the generator.

\textit{Format Reward ($R_{\text{format}}$).} 
This component serves as the prerequisite for parsing. We assign $R_{\text{format}}=1$ if the output $o_i$ strictly follows the XML protocol (containing \texttt{<relevance>}, \texttt{<analysis>}, and \texttt{<answer>} in order), and $0$ otherwise.

\textit{Accuracy Reward ($R_{\text{accuracy}}$).} 
To ensure correctness, we evaluate the final answer $\mathcal{O}$ against the ground truth. We employ normalized Exact Match (EM) over soft metrics to enforce precision. The reward is binary: $1$ for a match, $0$ for a mismatch.

\textit{Relevance Reward ($R_{\text{relevance}}$).} 
We evaluate the quality of the selected evidence $\mathcal{I}$ against the ground-truth reference set $\mathcal{G}$. Instead of a continuous score, we adopt a tiered scheme to enforce precision:
\begin{equation}
    R_{\text{relevance}} = 
    \begin{cases} 
    1 & \text{if } \mathcal{I} = \mathcal{G} \quad \text{(Perfect Match)}, \\
    0.5 & \text{if } \mathcal{I} \cap \mathcal{G} \neq \emptyset \text{ and } \mathcal{I} \neq \mathcal{G}, \\
    0 & \text{otherwise}.
    \end{cases}
\end{equation}
This design rewards the model for finding \textit{any} correct evidence but incentivizes it to find the \textit{exact} set of complete evidence.

\textit{Bonus Reward ($R_{\text{bonus}}$).} 
To amplify the distinction between ``acceptable'' and ``perfect'' generations, we introduce a sparse bonus signal. This reward is activated if and only if the model achieves perfect scores on all previous dimensions:
\begin{equation}
    R_{\text{bonus}} = \mathbb{I}(R_{\text{format}}=1 \land R_{\text{accuracy}}=1 \land R_{\text{relevance}}=1).
\end{equation}

To prioritize holistic alignment, we assign standard weights to the base metrics ($w_{\text{format}}=w_{\text{accuracy}}=w_{\text{relevance}}=1$) but assign a significantly higher weight to the bonus ($w_{\text{bonus}}=10$). This configuration serves as a strong gradient signal, encouraging the policy to converge towards outputs that are simultaneously structured, accurate, and well-grounded.

\subsection{Stabilized Policy Optimization via Unbiased Estimator}
\label{sec:kl_optimization}

Standard GRPO utilizes the $k_3$ estimator ($\hat{\mathbb{D}}_{KL}^{(k_3)} = \rho_t - \log \rho_t - 1$) to enforce KL constraints. While this estimator ensures non-negativity, it yields a biased gradient estimation when the policy deviates significantly from the reference. In this work, we adopt the $k_2$ estimator ($\hat{\mathbb{D}}_{KL}^{(k_2)} = \frac{1}{2}(\log \rho_t)^2$) to mitigate this issue.

Building on established derivations~\cite{rethinking-KL}, we compare the effective update directions of the true KL divergence against those derived from the $k_2$ and $k_3$ estimators. Let $\rho_t = \pi_{\text{ref}}(o_t)/\pi_\theta(o_t)$. The gradients with respect to $\theta$ are:
\begin{align}
    \text{True KL:\;\;} \quad -\nabla_\theta \mathbb{D}_{\mathrm{KL}} &= \mathbb{E} \left[ \log \rho_t \cdot \nabla_\theta \log \pi_\theta \right], \\
    \text{$k_2$ Estimator:} \quad -\nabla_\theta \hat{\mathbb{D}}_{KL}^{(k_2)} &= \mathbb{E} \left[ \log \rho_t \cdot \nabla_\theta \log \pi_\theta \right], \\
    \text{$k_3$ Estimator:} \quad -\nabla_\theta \hat{\mathbb{D}}_{KL}^{(k_3)} &= \mathbb{E} \left[ (\rho_t - 1) \cdot \nabla_\theta \log \pi_\theta \right].
\end{align}

As shown in Eq.~(13), the gradient of the $k_2$ estimator \textbf{strictly matches} the true KL gradient, confirming its property as an unbiased estimator. In contrast, the $k_3$ estimator in Eq.~(14) introduces a weighting factor of $(\rho_t - 1)$. While $(\rho_t - 1) \approx \log \rho_t$ holds when $\rho_t \approx 1$, this approximation degrades rapidly under large distribution shifts (where $\rho_t \gg 1$ or $\rho_t \ll 1$), leading to mis-scaled  updates.

Although the theoretical properties of $k_2$ have been discussed in the broader RL literature, our contribution lies in identifying its critical value specifically for RAG reasoning tasks. In such scenarios, the generator must explore complex reasoning paths that often diverge substantially from the backbone model (causing large $\rho_t$ fluctuations). By integrating the unbiased $k_2$ estimator, TRACE ensures global gradient fidelity and symmetric regularization. We empirically verify in our experiments that this substitution significantly enhances training stability and performance compared to the standard $k_3$ baseline.

\input{tables/main}

\section{Experiments}

\subsection{Experimental Setups}
\label{sec:experimental setups}

\input{tables/datasets}

\subsubsection{Datasets}

We evaluate our framework on three standard multi-hop Question Answering (QA) benchmarks: \textbf{HotpotQA}~\cite{yang2018hotpotqa}, \textbf{2WikiMultiHopQA} (2Wiki)~\cite{ho2020constructing}, and \textbf{MuSiQue}~\cite{trivedi2022musique}. 
These datasets characterize complex reasoning tasks where the questions are specifically designed hard to answer without consulting the specific retrieved documents, thereby preventing models from bypassing retrieval via parametric memory.
In their original configuration, each input consists of a question and a manually annotated context of 10--20 Wikipedia paragraphs. Among these, typically 2--4 paragraphs are labeled as \texttt{supporting\_facts} required to answer the question, while the others serve as hard distractors. 
The information across these retrieved paragraphs is designed to be non-overlapping which ensures that the target answer cannot be correctly derived from alternative retrieval paths. This property is vital for our training, as it guarantees that a correct reasoning process must strictly depend on the specific gold evidence, thereby validating our relevance reward mechanism.
To standardize these inputs for training, we adopt the preprocessing pipeline from PIKE-RAG~\cite{wang2025pike}, formatting each instance as a tuple $(q, \mathcal{C}, a^*, \mathcal{G})$. Here, $\mathcal{C}$ consists of the fixed candidate paragraphs, and $\mathcal{G}$ represents the set of ground-truth paragraph indices, which we construct by mapping the original sentence-level \texttt{supporting\_facts} to their corresponding paragraphs in $\mathcal{C}$.
Following prior practice~\cite{song2025r1}, we construct a composite training set of 25,000 examples (comprising 10,000 from HotpotQA, 10,000 from 2Wiki, and 5,000 from MuSiQue) and use a fixed set of 500 examples per dataset for evaluation, as detailed in Table~\ref{tab:dataset-detail}.

\subsubsection{Evaluation Metrics}
We assess model performance using three standard multi-hop QA metrics~\cite{yang2018hotpotqa, gao2023retrieval}: normalized \textbf{Exact Match} (EM), \textbf{F1 score} (F1), and \textbf{LLM-as-a-Judge} (LJ)~\cite{zheng2023judging}. EM and F1 measure surface-level string overlap with the gold answer, while LJ uses GPT-4o~\cite{gpt-4o} (version 2024-11-20) to provide a semantic judgment of answer correctness. In addition to these, we introduce two auxiliary evaluation metrics: \textbf{Format score} and \textbf{Relevance score}, which assess instruction adherence and citation accuracy. These scores follow the same evaluation criteria as used in the reward computation during training (see Section~\ref{sec:rewards}).

\subsubsection{Baselines}
We compare our method against a comprehensive set of baselines across three categories: \textbf{(1) Prompt-based models: }we include Qwen2.5-7B-Instruct~\cite{yang2024qwen2}, Llama3.1-8B-Instruct~\cite{grattafiori2024llama}, DeepSeek-R1-Distill-Qwen-7B~\cite{guo2025deepseek}, and Qwen3-8B~\cite{yang2025qwen3}, which directly answer questions with references\footnote{For reasoning-capable models, the internal thinking process is implicitly conducted during inference.}. We also consider SuRe~\cite{kim2024sure}, which enhances performance via summarized retrieval, and self-ask~\cite{press2022measuring}, which decomposes complex questions into sub-questions and answers them iteratively. \textbf{(2) SFT-based models: }we implement two variants: SFT-Direct, which directly generates the final answer, and SFT-Reasoning, which produces structured thoughts and answers. We also include SimpleDeepSearcher~\cite{simpledeepsearcher}, which trains models using automatically generated reasoning and retrieval trajectories.  \textbf{(3) RL-based models: }we implement a Naive-GRPO baseline using the original GRPO algorithm~\cite{shao2024deepseekmath}, where models are trained to output structured reasoning with format and accuracy rewards. In addition, we evaluate against four recent RL-based RAG frameworks: R1-Searcher~\cite{song2025r1}, Search-R1~\cite{searchr1}, ZeroSearch~\cite{zerosearch}, and ReSearch~\cite{research}, which all aim to teach the model to actively retrieve and reason through reinforcement learning. For consistency, we use their released checkpoints and replace their retrieved contexts with those provided by datasets. 

\subsubsection{Implementation details}
\label{sec:implementation}
We validate the TRACE framework on two blackbone models: Qwen2.5-7B-Instruct~\cite{yang2024qwen2} and Llama3.1-8B-Instruct~\cite{grattafiori2024llama}. Throughout the experimental section, we refer to the models trained with our framework as \textbf{TRACE-Qwen2.5-7B} and \textbf{TRACE-Llama3.1-8B}, respectively.
We train all models using the Open-R1~\cite{openr1} RL framework with DeepSpeed ZeRO Stage 2 optimization and bfloat16 mixed-precision. During inference, we concatenate all retrieved paragraphs provided by the dataset in their original order. The total batch size is 256, learning rate is set to 3e-6. We generate 7 rollout samples per input for reward estimation. We set temperature = 0.9 during rollout, KL coefficient $\beta = 0.04$, number of iterations per batch $\mu = 1$, and clipping parameter $\epsilon = 0.2$. 
All models are trained using 8 NVIDIA A100-80G GPUs, with 7 allocated for policy optimization and 1 dedicated to rollout inference via a vLLM~\cite{kwon2023efficient} engine. Our training takes about 20 hours.

\subsection{Main Results}
\label{sec:main_results}

As summarized in Table~\ref{tab:main_results}, \textbf{TRACE-Qwen2.5-7B} achieves the best performance across all datasets and metrics, consistently outperforming all baselines of similar scale, particularly on  2Wiki where it achieves an Exact Match (EM) gain of over 30\% (from 33.4\% to 66.0\%). The improvements observed in \textbf{TRACE-Llama3.1-8B} validate the effectiveness of our framework across backbones. Furthermore, our results reveal that RL-based models significantly outperform SFT-based and prompt-based methods; this disparity underscores that for multi-hop QA tasks lacking high-quality reasoning annotations, the self-exploration mechanism of RL is more effective at refining reasoning logic and evidence utilization than imitation learning.

\input{tables/closed_comparison}

To contextualize our performance within the broader landscape, we compare TRACE-Qwen2.5-7B against leading commercial models on HotpotQA in Table~\ref{tab:closed_comparison}.
Remarkably, our 7B model achieves an EM score of 62.8\%, effectively matching GPT-4o~\cite{gpt-4o} (62.8\%) and outperforming DeepSeek-R1~\cite{song2025r1} (61.8\%) in this specific setting. 
While a gap remains compared to OpenAI o1~\cite{openai2024reasoning} (65.6\%), TRACE significantly narrows this gap.
These results highlight a critical finding: by enforcing structured evidence navigation and stabilizing RL optimization, small open-source models can achieve reasoning proficiency comparable to commercial models.

\input{tables/transparency_results}

A core objective of TRACE is enhancing interpretability via explicit evidence traceability. As shown in Table~\ref{tab:transparency_results}, TRACE-Qwen2.5-7B shows perfect Format score, resolving the structural instability often found in backbone models. This structural adherence is accompanied by a huge leap in Relevance score, with absolute improvements of 23.0\% on HotpotQA and 23.5\% on 2Wiki. Such significant gains confirm that our framework does not merely mimic the format but truely learns to identify and ground its reasoning in the correct supporting facts, fulfilling transparent RAG generator.

\input{tables/dataset_generalization}

\begin{figure*}[t]
    \centering
    \includegraphics[width=1\textwidth]{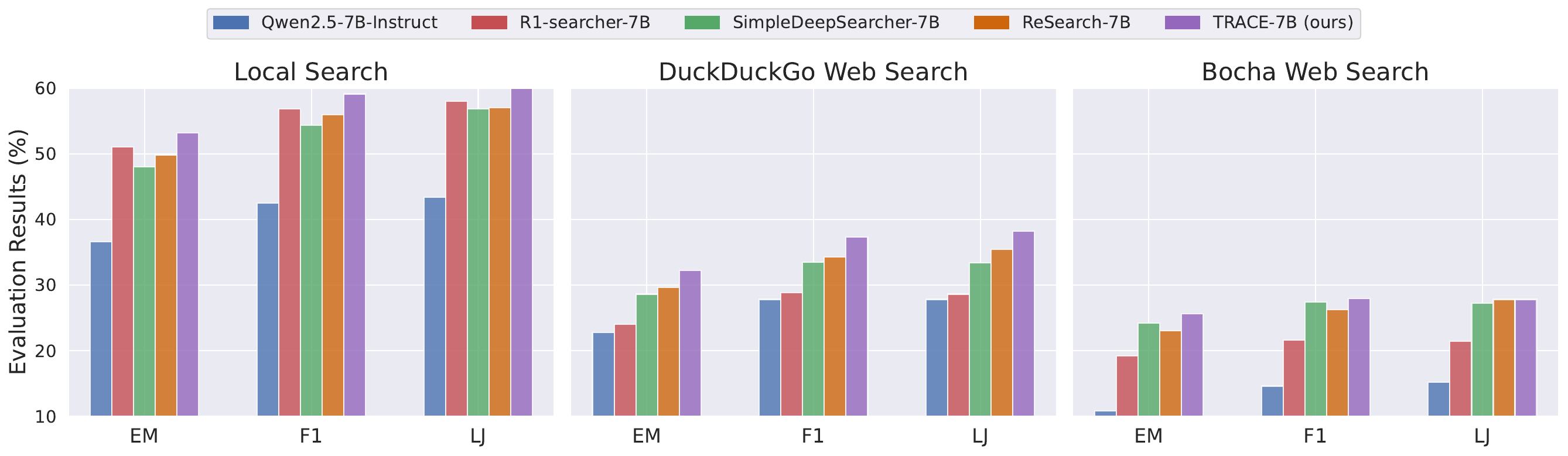}
    \caption{
    Performance comparison on 2Wiki under three distinct retrieval settings: (1) Local Search (Wikipedia), (2) DuckDuckGo Web Search, and (3) Bocha Web Search. \textbf{TRACE-Qwen2.5-7B} consistently outperforms both SFT and other RL baselines across all metrics, demonstrating robustness against the variable noise levels inherent in real-world retrieval.
    }
    \label{fig:retrieval_generalization}
\end{figure*}

\subsection{Generalization Ability}
\label{sec:generalization}

We assess the generalization ability of \textbf{TRACE} by evaluating its performance across unseen domains and varying retrieval conditions to ensure the learned policy is transferable and effective under different levels of external noise.

\noindent\textbf{Cross-Domain Generalization.} 
To verify that TRACE learns transferable reasoning skills, we conduct zero-shot cross-domain experiments. As shown in Table~\ref{tab:dataset_generalization}, we train the model on subsets of the training data and evaluate it on the held-out datasets. These results, which closely approach the performance of the model trained on the full dataset mixture (\textbf{HWM}), indicate that our RL optimization facilitate the learning of universal evidence-grounding patterns that generalize well across diverse reasoning contexts.

\noindent\textbf{Retrieval setting generalization.} 
Real-world RAG deployment requires models to handle retrieval results from various sources. We evaluate TRACE-Qwen2.5-7B on 2Wiki under three distinct retrieval settings: (1) \textbf{Local Search} based on Wikipedia~\cite{petroni2020kilt}, (2) \textbf{DuckDuckGo Web Search}~\cite{hands2012duckduckgo}, and (3) \textbf{Bocha Web Search}~\cite{bochaai}. As illustrated in Figure~\ref{fig:retrieval_generalization}, despite containing significant noise and irrelevant contexts, TRACE-Qwen2.5-7B consistently outperforms both SFT and RL-based baselines across all scenarios. This demonstrates that our framework successfully internalizes the ability to filter out noisy information and accurately identify high-value evidence, a capability critical for stable real-world RAG deployment.

\begin{figure}[t]
    \centering
    \includegraphics[width=0.8\columnwidth]{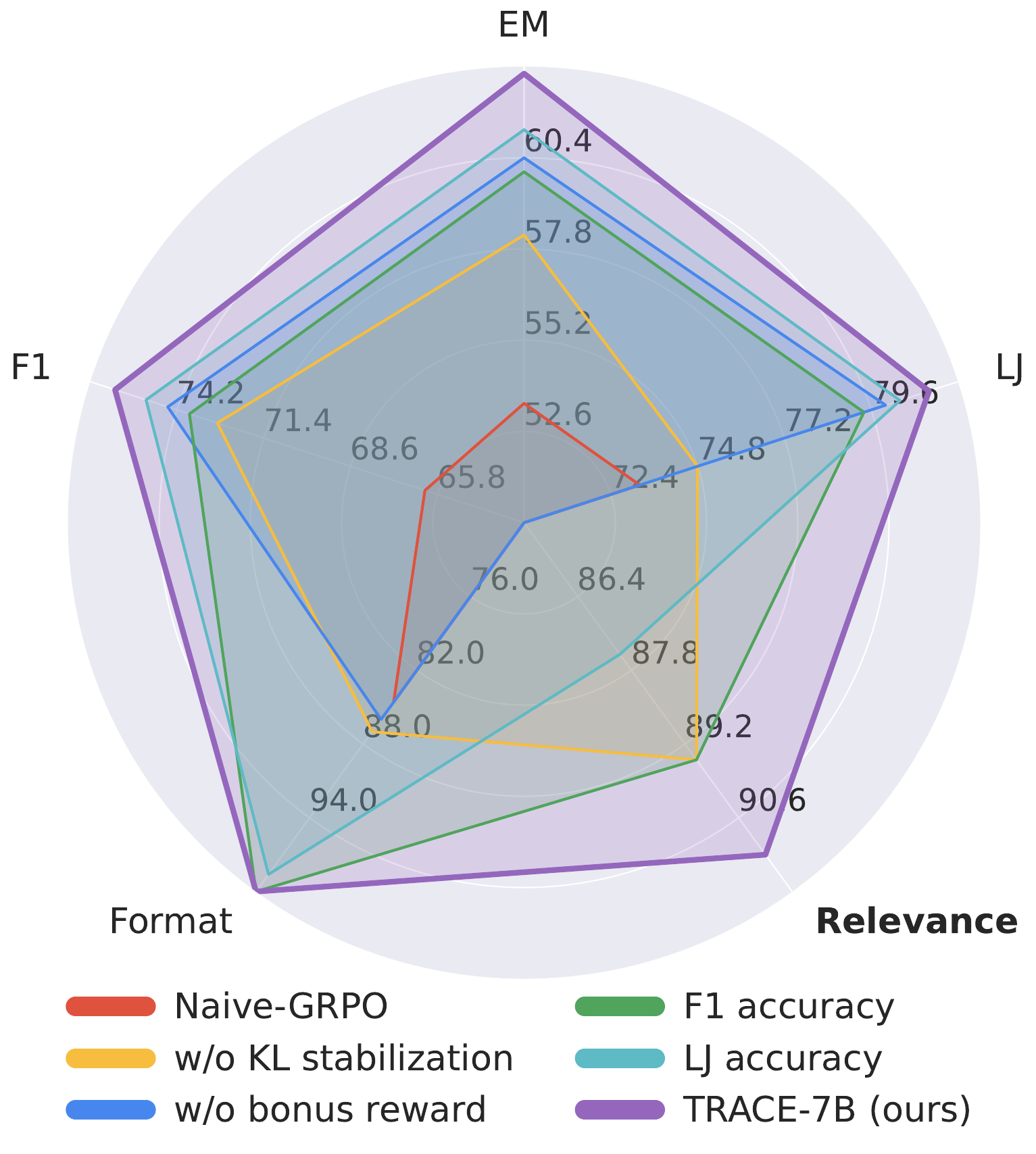}
    \caption{
    Radar chart comparing six variants across five evaluation metrics on HotPotQA. TRACE shows consistent improvements in output quality and interpretability.
    }
    \label{fig:ablation-radar}
\end{figure}

\input{tables/bonus_ablation}

\subsection{Ablation Study and Stability Analysis}
\label{sec:ablation}

\begin{figure}[t]
    \centering
    \includegraphics[width=0.9\columnwidth]{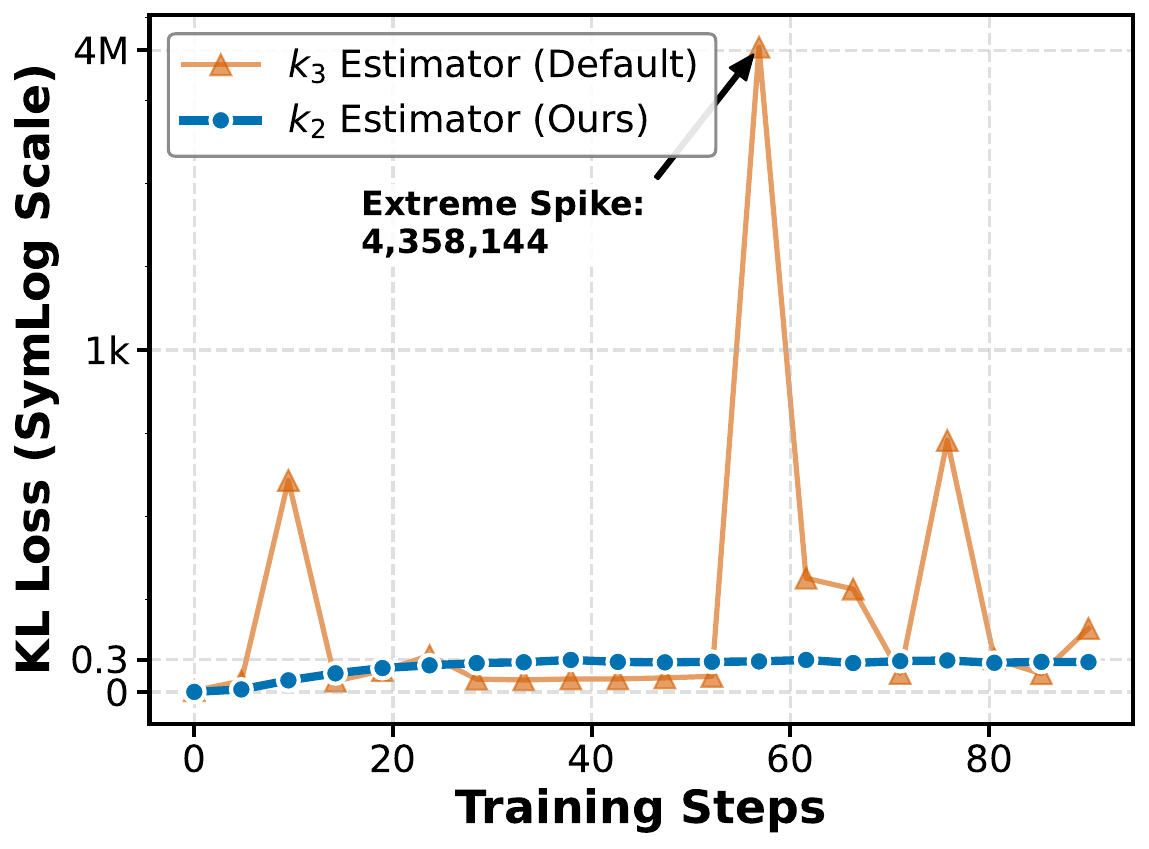}
    \caption{
    Comparison of KL estimators during training. 
    }
    \label{fig:kl-stability}
\end{figure}

To validate the contribution of each component within TRACE, we conduct a comprehensive ablation study on HotpotQA. Figure~\ref{fig:ablation-radar} compares the TRACE-Qwen2.5-7B against five variants: (1) \textbf{Naive-GRPO}, which uses basic GRPO without our structured protocol or specific rewards; (2) a variant \textbf{without KL stabilization}, utilizing the standard $k_3$ estimator; (3) a variant \textbf{without the bonus reward}; and (4) variants using \textbf{soft accuracy rewards} such as F1 or LJ.

The radar chart in Figure~\ref{fig:ablation-radar} illustrates that TRACE achieves the best performance across all dimensions. The poor performance of Naive-GRPO on transparency metrics confirms that RL alone is insufficient for spontaneously developing structured attribution capabilities without the explicit constraints provided by our protocol. Furthermore, variants utilizing soft rewards (F1/LJ Accuracy) underperform compared to our model, this suggest that strict binary rewards provide clearer and more definitive signals for optimizing the reasoning logic in multi-hop QA tasks, whereas softer metrics may introduce ambiguity into the policy update.

The critical role of the holistic bonus reward is further evidenced by the training trajectory detailed in Table~\ref{tab:bonus_ablation}. When bonus reward is removed, we observe a significant alignment trade-off during the optimization process. Although the model continues to refine its answer accuracy, its adherence to the transparency protocol suffers a catastrophic collapse, evidenced by the relevance score dropping from 86.2\% to 13.6\%. This proves that bonus reward acts as an essential stabilizer preventing the policy from taking shortcuts.

We also analyze the impact of the $k_2$ estimator on training stability. As shown in Figure~\ref{fig:kl-stability}, the standard $k_3$ estimator exhibits extreme numerical instability, with gradient spikes reaching magnitudes as high as $4.3 \times 10^6$ during the middle stages of training. Such spikes often lead to policy divergence or the generation of degenerate, repetitive sequences. In contrast, the $k_2$ estimator maintains a smooth and bounded KL divergence throughout the entire optimization process. This stability is vital for our framework, as learning a rigid structured protocol requires consistent and reliable policy updates. The noticeable performance drop in the variant without KL stabilization shown in Figure ~\ref{fig:ablation-radar} confirms that numerical robustness directly translates into superior end-task performance.

\begin{figure}[t]
    \centering
    \includegraphics[width=\columnwidth,
        trim=7mm 7mm 7mm 7mm
    ]{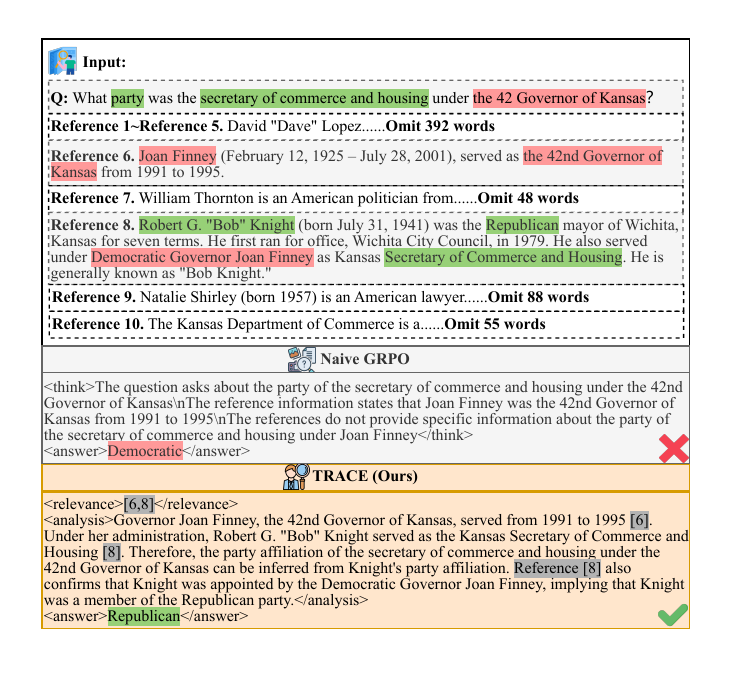}
    \caption{
    A Case Study of Naive GRPO and TRACE-Qwen2.5-7B. TRACE correctly extracts relevant paragraphs, provides a structured explanation, and delivers the right answer.
    }
    \label{fig:reasoning-case}
\end{figure}

\subsection{Case Study}
\label{sec:case_study}

Figure~\ref{fig:reasoning-case} illustrates a representative case study from HotpotQA. For a multi-hop query regarding the political party of an official serving under the 42nd Governor of Kansas, the naive GRPO baseline produces a structured response that is nonetheless logically incorrect. Due to the lack of explicit evidence grounding, the baseline falls into a distractor trap, mistakenly attributing the Governor’s political party (Democratic) to the Secretary mentioned in the same paragraph. In contrast, TRACE first identifies the correct supporting documents [6, 8] and utilizes this selection to anchor its reasoning process. By clearly distinguishing between the entities and their respective attributes across the reasoning chain, TRACE correctly identifies the official as a Republican and delivers the right answer. This comparison demonstrates that the transparent protocol in TRACE does not merely provide an auditable trail, but actively prevents reasoning hallucinations by compelling the model to verify each logical link against specific, pre-selected evidence.

\section{Conclusion}
\label{sec:conclusion}

In this work, we introduced \textbf{TRACE} (\textbf{T}ransparent \textbf{RA}G with eviden\textbf{CE} tracing), a RL framework that demonstrates how improve the performance of RAG generator through clear evidence traceability. Moving beyond traditional systems that often treat attribution as a post-hoc annotation, TRACE enforces a structured protocol that requires models to explicitly anchor each reasoning step in verified references prior to deriving the final answer. To facilitate the robust learning of such complex behaviors, we developed an adaptive reward mechanism with a holistic alignment bonus and a stabilized, gradient-unbiased KL estimator. This unified optimization strategy effectively mitigates the reward sparsity and training instability often encountered in constrained RAG reasoning. 
Extensive evaluations across multiple multi-hop benchmarks show that TRACE delivers substantial accuracy gains of 10--30\%, effectively approaching the reasoning ability of advanced commercial models. By harmonizing verifiable traceability with superior performance, TRACE offers a robust path toward trustworthy deployment in high-stakes domains.
Looking ahead, we hope this work inspires the community to look beyond mere answer correctness and place equal emphasis on the rigorous examination of reasoning traces, fostering a future where RAG systems are defined not only by what they know, but by how transparently they think.

\bibliographystyle{ACM-Reference-Format}
\bibliography{ref}

\end{document}

%% file: tables/main.tex
\begin{table*}[ht]
\centering
\caption{
Main experimental results across three datasets and three evaluation metrics. Models are grouped by training methodology. The last three columns report the \textbf{Average} performance across all datasets. \textbf{Bold} indicates the best score; \underline{underline} indicates the second-best. Metrics: EM = Exact Match(\%), F1 = F1 score(\%), LJ = LLM-as-a-Judge(\%).
}
\begin{tabular}{lccc|ccc|ccc|ccc}
\toprule
\multicolumn{1}{c}{\multirow{2}{*}{\normalsize\textbf{Model}}} & 
\multicolumn{3}{c}{\textbf{HotpotQA}} &
\multicolumn{3}{c}{\textbf{2Wiki}} &
\multicolumn{3}{c}{\textbf{MuSiQue}} & 
\multicolumn{3}{c}{\textbf{Average}}
\\
\cmidrule(lr){2-4} \cmidrule(lr){5-7} \cmidrule(lr){8-10} \cmidrule(lr){11-13} 
& EM & F1 & LJ & EM & F1 & LJ & EM & F1 & LJ & EM & F1 & LJ \\
\midrule

\multicolumn{13}{l}{\textit{\textbf{Prompt-based models}}} \\
Llama3.1-8B-Instruct & 52.8 & 67.6 & 73.8 & 39.8 & 47.8 & 48.4 & 24.8 & 37.0 & 32.4 & 39.1 & 50.8 & 51.5 \\
Qwen2.5-7B-Instruct & 48.4 & 62.8 & 66.0 & 33.4 & 42.4 & 41.2 & 25.2 & 35.4 & 30.6 & 35.7 & 46.9 & 45.9 \\
DeepSeek-R1-Distill-Qwen-7B & 33.2 & 48.7 & 71.2 & 29.0 & 40.7 & 65.8 & 11.6 & 18.4 & 27.8 & 24.6 & 35.9 & 54.9 \\
Qwen3-8B & 58.2 & 71.9 & 76.8 & \underline{65.2} & \underline{72.7} & \underline{78.4} & 33.6 & 39.7 & 39.4 & \underline{52.3} & 61.4 & 64.9 \\
SuRe-GPT-4o & 49.0 & 69.5 & 74.2 & 49.0 & 60.7 & 63.2 & 19.4 & 30.7 & 32.4 & 39.1 & 53.6 & 56.6 \\
SuRe-Qwen-7B & 50.4 & 64.8 & 69.6 & 36.4 & 44.8 & 44.0 & 14.6 & 23.3 & 22.6 & 33.8 & 44.3 & 45.4 \\
self-ask-GPT-4o & 40.8 & 57.7 & 66.2 & 43.0 & 56.5 & 61.4 & 16.2 & 30.5 & 34.6 & 33.3 & 48.2 & 54.1 \\
\addlinespace

\midrule

\multicolumn{13}{l}{\textit{\textbf{SFT-based models}}} \\
Qwen-7B-SFT-direct & 49.8 & 62.8 & 66.2 & 54.0 & 60.9 & 62.0 & 16.8 & 25.4 & 21.4 & 40.2 & 49.7 & 49.9 \\
Qwen-7B-SFT-reasoning & 40.0 & 53.2 & 59.9 & 53.6 & 61.5 & 62.2 & 11.2 & 17.2 & 17.4 & 34.9 & 44.0 & 46.5 \\
SimpleDeepSearcher-7B & 50.0 & 63.3 & 68.6 & 63.2 & 70.7 & 74.8 & 25.8 & 34.7 & 34.4 & 46.3 & 56.2 & 59.3 \\
\addlinespace

\midrule

\multicolumn{13}{l}{\textit{\textbf{RL-based models}}} \\
Naive-GRPO & 53.4 & 66.2 & 73.2 & 62.0 & 68.4 & 72.0 & 33.2 & 43.5 & 42.0 & 49.5 & 59.4 & 62.4 \\
R1-Searcher-7B & \underline{59.0} & \underline{73.1} & \underline{79.1} & 63.2 & 70.4 & 73.0 & 25.0 & 37.0 & 32.7 & 49.1 & 60.2 & 61.6 \\
Search-R1-7B & 53.2 & 68.5 & 76.6 & 51.8 & 61.7 & 66.0 & 30.6 & 40.8 & 43.2 & 45.2 & 57.0 & 61.9 \\
ZeroSearch-7B & 51.8 & 64.6 & 70.6 & 43.6 & 51.7 & 53.2 & 27.0 & 36.4 & 34.2 & 40.8 & 50.9 & 52.7 \\
ReSearch-7B & 57.8 & 72.9 & 77.8 & 58.6 & 65.8 & 68.4 & 30.0 & 39.8 & 40.7 & 48.8 & 59.5 & 62.3 \\
\addlinespace

\midrule

\multicolumn{13}{l}{\textit{\textbf{Ours}}} \\
\textbf{TRACE-Llama3.1-8B} & 55.2 & 70.3 & 78.0 & 62.2 & 71.0 & 73.5 & \underline{35.8} & \underline{44.5} & \underline{46.6} & 51.1 & \underline{61.9} & \underline{66.0} \\
\textbf{TRACE-Qwen2.5-7B} & \textbf{62.8} & \textbf{76.2} & \textbf{81.2} & \textbf{66.0} & \textbf{75.2} & \textbf{78.6} & \textbf{40.0} & \textbf{52.0} & \textbf{50.8} & \textbf{56.3} & \textbf{67.8} & \textbf{70.2} \\
\bottomrule
\end{tabular}
\label{tab:main_results}
\end{table*}

%% file: tables/datasets.tex
\begin{table}[ht]
\centering
\caption{Detailed dataset statistics used in our experiments. Hop counts are derived from the \texttt{supporting\_facts} fields.}
\begin{tabular}{lccccccccc}
\toprule
\textbf{Dataset} & \textbf{Split} & \textbf{Data Size} & \textbf{\# Para} & \textbf{Avg. Hops} \\
\midrule
HotpotQA & Train & 10,000 & 10 & 2.00 \\
HotpotQA & Test  & 500 & 10 & 2.00 \\
2Wiki & Train & 10,000 & 10 & 2.39 \\
2Wiki & Test  & 500 & 10 & 2.19 \\
MuSiQue & Train & 5,000 & 20 & 2.85 \\
MuSiQue & Test  & 500 & 20 & 2.73 \\
\midrule
\textbf{All Train} & -- & 25,000 & -- & 2.30 \\
\bottomrule
\end{tabular}
\label{tab:dataset-detail}
\end{table}

%% file: tables/closed_comparison.tex
\begin{table}[t]
\centering
\caption{
Performance comparison between our \textbf{TRACE-Qwen2.5-7B} and leading reasoning models on HotpotQA.
}
\begin{tabular}{lccc}
\toprule
\textbf{Model} & \textbf{EM} & \textbf{F1} & \textbf{LJ} \\
\midrule
GPT-4o         & 62.8 & 78.8 & 82.0 \\
DeepSeek-R1    & 61.8 & 78.1 & 83.6 \\
OpenAI o1      & \textbf{65.6} & \textbf{81.6} & \textbf{89.0} \\
\midrule
\textbf{TRACE-Qwen2.5-7B (Ours)}  & 62.8 & 76.2 & 81.2 \\
\bottomrule
\end{tabular}
\label{tab:closed_comparison}
\end{table}

%% file: tables/transparency_results.tex
\begin{table}[t]
\centering
\caption{
Evaluation of transparency metrics: \textbf{Format} (Fmt.) score and \textbf{Relevance} (Rel.) score. TRACE-Qwen2.5-7B achieves perfect protocol adherence and significantly higher evidence grounding accuracy.
}
\small
\setlength{\tabcolsep}{3.5pt}
\begin{tabular}{lcc|cc|cc}
\toprule
\multicolumn{1}{c}{\multirow{2}{*}{\textbf{Model}}} &
\multicolumn{2}{c}{\textbf{HotpotQA}} &
\multicolumn{2}{c}{\textbf{2Wiki}} &
\multicolumn{2}{c}{\textbf{MuSiQue}} \\
\cmidrule(lr){2-3} \cmidrule(lr){4-5} \cmidrule(lr){6-7}
& Fmt. & Rel. & Fmt. & Rel.  & Fmt. & Rel. \\
\midrule
Qwen2.5-7B-Instruct & 91.6 & 68.3 & 97.8 & 67.3 & 80.2 & 48.4 \\
\textbf{TRACE-Qwen2.5-7B} & \textbf{100.0} & \textbf{91.3} & \textbf{100.0} & \textbf{90.8} & \textbf{100.0} & \textbf{59.5} \\
\bottomrule
\end{tabular}
\label{tab:transparency_results}
\end{table}

%% file: tables/dataset_generalization.tex
\begin{table}[t]
\centering
\caption{
Zero-shot cross-domain generalization evaluation. We train \textbf{TRACE-Qwen2.5-7B} on subsets of data (e.g., ``WM'' denotes training on 2\textbf{W}iki + \textbf{M}uSiQue) and evaluate on all test sets, including the held-out dataset.
}
\small
\setlength{\tabcolsep}{4pt}
\begin{tabular}{lccc|ccc|ccc}
\toprule
\multicolumn{1}{c}{\multirow{2}{*}{\normalsize\textbf{Trainset}}} &
\multicolumn{3}{c}{\textbf{HotpotQA}} &
\multicolumn{3}{c}{\textbf{2Wiki}} &
\multicolumn{3}{c}{\textbf{MuSiQue}}
\\
\cmidrule(lr){2-4} \cmidrule(lr){5-7} \cmidrule(lr){8-10}
& EM & F1 & LJ & EM & F1 & LJ & EM & F1 & LJ \\
\midrule
WM & 56.8 & 71.1 & 76.4 & 64.2 & 73.2 & 73.4 & 39.4 & 50.4 & 48.6 \\
HM & 59.8 & 74.1 & 79.6 & 57.8 & 67.1 & 68.0 & \textbf{40.4} & 51.9 & 47.8  \\
\textbf{HWM} & \textbf{62.8} & \textbf{76.2} & \textbf{81.2} & \textbf{66.0} & \textbf{75.2} & \textbf{77.4} & 40.0 & \textbf{52.0} & \textbf{50.8} \\
\bottomrule
\end{tabular}
\label{tab:dataset_generalization}
\end{table}

%% file: tables/bonus_ablation.tex
\begin{table}[t]
\centering
\caption{
Training trajectory of TRACE without the bonus reward on HotpotQA. While accuracy (EM) improves, the structural integrity (Format/Relevance) degrades significantly over time, confirming the role of the bonus as a stabilizer.
}
\begin{tabular}{lcccc}
\toprule
\textbf{Training Step} & \textbf{EM} & \textbf{F1} & \textbf{Format} & \textbf{Relevance} \\
\midrule
Checkpoint-25 & 58.4 & 73.2 & \textbf{99.4} & \textbf{86.2} \\
Checkpoint-50 & 58.8 & 73.2 & 95.6 & 81.9 \\
Checkpoint-75 & 58.2 & 73.2 & 91.0 & 29.5 \\
Final (Step 95) & \textbf{60.4} & \textbf{74.5} & 86.0 & 13.6 \\
\bottomrule
\end{tabular}
\label{tab:bonus_ablation}
\end{table}